\documentclass[10pt,twocolumn,letterpaper]{article}

\usepackage{cvpr}              

\usepackage[dvipsnames]{xcolor}

\definecolor{cvprblue}{rgb}{0.21,0.49,0.74}
\usepackage[pagebackref,breaklinks,colorlinks,citecolor=cvprblue]{hyperref}
\usepackage{multirow}
\usepackage{makecell}
\usepackage{diagbox}
\def\paperID{9470} 
\def\confName{CVPR}
\def\confYear{2024}

\title{Control4D: Efficient 4D Portrait  Editing with Text }

\author{Ruizhi Shao$^{1}$, Jingxiang Sun$^{1}$, Cheng Peng$^{1}$, Zerong Zheng$^{1,2}$,  Boyao Zhou$^{1}$, Hongwen Zhang$^{1}$, Yebin Liu$^{1}$\\
$^{1}$Department of Automation, Tsinghua University $^{2}$NNKosmos Technology 
}

\begin{document}

\twocolumn[{
\maketitle
\begin{center}
    \captionsetup{type=figure}
    \vspace{-8mm}
    \includegraphics[width=1.\textwidth]{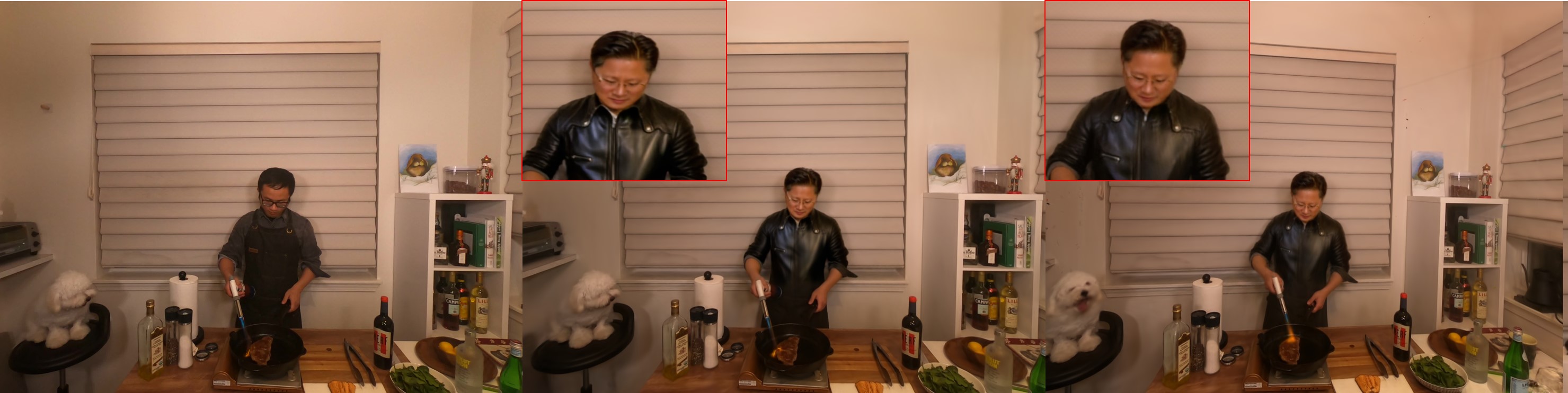}
    \vspace{-6mm}
    \captionof{figure}{We propose Control4D, an approach to high-fidelity and spatiotemporal-consistent 4D portrait editing with only text instructions. Given the multi-view videos as shown in the left and text instructions "Jensen Huang is roasting steak", Control4D generates realistic and 4D consistent editing results presented in the middle and right.}
\end{center}
}]

\begin{abstract}
We introduce Control4D, an innovative framework for editing dynamic 4D portraits using text instructions. Our method addresses the prevalent challenges in 4D editing, notably the inefficiencies of existing 4D representations and the inconsistent editing effect caused by diffusion-based editors. We first propose GaussianPlanes, a novel 4D representation that makes Gaussian Splatting more structured by applying plane-based decomposition in 3D space and time. This enhances both efficiency and robustness in 4D editing. Furthermore, we propose to leverage a 4D generator to learn a more continuous generation space from inconsistent edited images produced by the diffusion-based editor, which effectively improves the consistency and quality of 4D editing.
Comprehensive evaluation demonstrates the superiority of Control4D, including significantly reduced training time, high-quality rendering, and spatial-temporal consistency in 4D portrait editing. The link to our project website is: \href{https://control4darxiv.github.io/}{https://control4darxiv.github.io/}.

\end{abstract}


\section{Introduction} 
The realm of 4D scene reconstruction has witnessed advancements with the advent of dynamic neural 3D representation~\cite{Pumarola20, Park21a, Shao22, Fridovich23, Liu}. These innovations have significantly enhanced our ability to capture and represent dynamic scenes. Despite these advances, the interactive editing of these 4D scenes still poses substantial challenges. The primary challenge involves ensuring both spatial-temporal consistency and high quality in 4D editing.

Available 4D editing techniques~\cite{jiang20234d, qiao2022neuphysics}, while effective for fundamental tasks like object removal or color modification, often fall short in delivering interactive and advanced editing functionalities. Recently, a groundbreaking framework based on text-to-image (T2I) diffusion model~\cite{Poole} has emerged for 3D generation and editing. It integrates a neural 3D representation such as NeRF with an image diffusion model and achieves text-to-3D generation~\cite{Poole, Magic3D, Metzer, chen2023fantasia3d} or editing~\cite{Instruct-NeRF2NeRF} by iteratively aligning images rendered from the 3D representation with those generated by the diffusion model. This diffusion-based framework allows for more flexible and enhanced editing through textual control.

Building on this framework, a straightforward approach to 4D editing involves transitioning from a 3D to a 4D representation. However, it faces two primary challenges: First, 4D representations such as dynamic NeRFs require dense sampling along the rays to render images, which is slow and highly memory-intensive~\cite{Pumarola20, Shao22, Fridovich23}. Such inefficiency significantly increases the time required for editing in 4D scenarios. On the other hand, current T2I diffusion models lack consistency in editing different images~\cite{Instruct-NeRF2NeRF}. This inconsistency is more apparent in 4D editing, as the results vary across different spatial perspectives and over time, making 4D editing extremely challenging.

In this paper, we address these challenges and present Control4D, a novel method for efficient, high-quality, and consistent 4D dynamic portrait editing with text as input. Firstly, to enhance the efficiency of 4D representation, we propose to extend an explicit 3D representation, Gaussian Splatting, to a 4D dynamic representation. Gaussian Splatting is an emerging representation that has demonstrated its efficiency in training and rendering for 3D reconstruction~\cite{kerbl20233d} and generation~\cite{tang2023dreamgaussian}. However, as it uses discrete Gaussian point clouds where every point is independent from each other, it easily introduces noise during the 4D editing process, where the edited images are not consistent in both space and time. To address this issue, we first propose to construct the spatial structure to describe the attributes of discrete Gaussian points by a unified, structured tri-plane~\cite{chan2022efficient} representation. Specifically, we project each Gaussian point onto three feature planes and employ an MLP to integrate features and derive their attributes, which not only ensures efficiency but also enhances robustness. Then, we extend Gaussian Splatting to 4D by defining a canonical Gaussian point cloud and allowing each point to move with time. To regularize the flow of discrete points, we also project their positions with time into 9 planes~\cite{Shao22} to make the flow more structured. With the tri-planar structure for the canonical space and the 4D plane-based structure for the 4D flow, we introduce GaussianPlanes representation, which significantly reduces the time cost and improves spatiotemporal consistency in 4D editing. 

Although GaussianPlanes significantly improves the efficiency of representation, implementing 4D editing based on it still encounters a bottleneck. This bottleneck lies in the T2I diffusion model, as the diffusion-based editor adopts a 2D generation process and produces inconsistent edits in 4D space across time and viewpoints. Consequently, when optimized with these inconsistent images, the dynamic scene model tends to diverge or produce blurry and smoothed outcomes. To overcome this challenge, we propose a 4D generator to mitigate the issue of inconsistent supervision arising from the edited dataset. The key insight of our method is to learn a more continuous GAN latent space based on the edited images produced by the editor, avoiding direct but inconsistent supervision. Specifically, we introduce additional latent properties to GaussianPlanes and incorporate it with a 2D super-resolution module, constructing a 4D generator, capable of producing high-resolution images based on the rendered latent features. Simultaneously, we employ a discriminator to learn the generation distribution from the edited images, which then provides discrimination signals for updating the generator. To ensure stable training, we extract multi-level information from the edited images and utilize it to facilitate the generator's learning process.

We conduct comprehensive evaluation of our approach using a diverse collection of dynamic portraits. To validate the efficacy of our design, we conduct ablation studies and compare our method with a 4D extension of InstructNeRF2NeRF~\cite{Instruct-NeRF2NeRF}. The evaluation demonstrates the efficiency and remarkable capabilities of our method in achieving both photo-realistic rendering and spatio-temporal consistency in 4D portrait editing.
To sum up, our main contributions are listed as follows:

\begin{itemize}[leftmargin=*]
    \item We propose an efficient and robust 4D representation GaussianPlanes for 4D editing by applying plane-based decomposition to structure Gassian Splatting in both space and time. 
    \item We introduce a 4D generator to learn from the 2D diffusion-based editor, which reduces the effect of inconsistent supervision signals and enhances the quality of 4D editing.
    \item Building upon the proposed GaussianPlanes and 4D generator, We introduce Control4D, a novel framework for flexible 4D portrait editing with text, which
 significantly reduces the training time, achieves high-quality rendering, and ensures spatio-temporal consistency.
\end{itemize}

\section{Related Work}

\subsection{2D Diffusion Models}

Diffusion models iteratively transform random samples into ones resembling target data~\cite{Sohl15,Ho20,kingma2021variational, dhariwal2021diffusion}. Enhanced with pre-trained models~\cite{CLIP}, they solve multi-modal tasks like text-to-image generation~\cite{GLIDE,Ho_Salimans,DALLE2}. VQdiffusion\cite{Gu21} and LDMs \cite{Rombach22} bolster performance by operating within an autoencoder's latent space. Although these models have found success, temporally inconsistent issues emerge in videos and 4D scenes.

Research has also concentrated on diffusion-based video generation and editing. Video Diffusion Models (VDM)\cite{Ho22} use U-Net architecture to train image and video data jointly, while approaches like ImagenVideo\cite{IMAGENVIDEO} enable high-resolution video generation. Various methods aim to transfer text-image generation to text-video, but due to training costs, many focus on text-prompted video editing~\cite{Singer22,Zhou22,ceylan2023pix2video,khachatryan2023text2video,Bar22,Esser,Wu22, li2023sweetdreamer}. These efforts underscore the potential of text-based video editing, yet challenges related to temporal consistency, quality generation, and viewpoint alterations persist.

\subsection{NeRF-Based 3D Generation and Editing}

NeRFs~\cite{Mildenhall20} have gained widespread popularity for producing realistic 3D scene reconstruction and novel views based on calibrated photographs, and have been further developed in numerous subsequent studies~\cite{Tewari21}. Nevertheless, NeRFs still pose a challenge for editing purposes, primarily due to their underlying representation. 

NERF editing researchers have focused on utilizing GANs~\cite{goodfellow2020generative} and Diffusion models for their powerful generative capabilities. GAN-based methods have seen a proliferation of novel architectures that combine implicit or explicit 3D representations with neural rendering techniques, achieving promising results~\cite{nguyen2019hologan,nguyen2020blockgan,wu2016learning,zhu2018visual,gu2021stylenerf, chan2022efficient}. However, voxel-based GANs face challenges such as high memory requirements and computational burden when training high-resolution 3D GANs. On the other hand, diffusion-based methods have two primary approaches for extending 2D editing to 3D NeRFs. The first involves using Stable Diffusion with score distillation sampling (SDS) loss to generate 3D NeRFs using the 2D diffusion-prior, as seen in DreamFusion~\cite{Poole} and its follow-ups~\cite{tang2023make, Wang22, Magic3D, shen2023anything,li20233d, chen2023fantasia3d, cao2023dreamavatar, seo2023ditto, po2023compositional, singer2023text, liu2023zero1to3, wang2023prolificdreamer, shi2023mvdream, huang2023humannorm, sun2023dreamcraft3d, tang2023dreamgaussian}. However, these methods can only generate isolated objects lacking fine-level control over synthesized outputs. The second approach utilizes dataset update(DU) to guide NeRF convergence iteratively, as seen in Instruct-NeRF2NeRF~\cite{Instruct-NeRF2NeRF}, but it has network convergence issues and can be cost-intensive.

\subsection{NeRF for Dynamic Scenes}
To expand the success of NeRF into the temporal domain, researchers have pursued the strategy of modeling scenes in 4D domain with time dimension. DyNeRF~\cite{Li21} proposes a keyframe-based training strategy to extend NeRF with time-conditioning. VideoNeRF \cite{Xian21} learns a spatiotemporal irradiance field directly from a single video and resolves the shape-motion ambiguities in monocular inputs by incorporating depth estimation. Meanwhile, NeRFlow \cite{Du20} and DCT-NeRF\cite{Wang21} utilize point trajectories to regularize network optimization. Park et al.\cite{Park21a,Park21b}; Pumarola et al. \cite{Pumarola20}; Tretschk et al. \cite{Tretschk21} adopt a similar framework that introduce a separate MLP to predict scene deformations for multi-view and monocular videos, respectively.Another approach for dynamic scenes is DeVRF \cite{Liu}, which adopts a voxel-based representation to model both the 3D canonical space and the 4D deformation field. Additionally, methods including Neuralbody~\cite{Peng20} and ~\cite{Weng, zheng2022structured, liu2021neural, zheng2023avatarrex} leverage parametric body templates as semantic priors to achieve photo-realistic novel view synthesis of complex human performances. Recently, to achieve higher quality with lower memory, NeRFPlayer \cite{Song} have decomposed the 4D space into regions of static, deforming, and newly appeared content. Meanwhile, more compact and efficient representations, such as \cite{Fridovich23, Cao_Johnson, xu20234k4d, HumanRF} are proposed, significantly boosting the rendering quality and efficiency.

\subsection{Gaussian Splatting}
3D Gaussian Splatting (3DGS)~\cite{kerbl20233d} offers a high-quality, swift alternative to Neural Radiance Fields (NeRF), leveraging differentiable 3D Gaussians for efficient rasterization. Unlike NeRF and other implicit 3D representations~\cite{wang2021neus,niemeyer2020differentiable} which render images based on volume rendering, 3D-GS employs a splatting method for image rendering, resulting in real-time speed. The successor, 4D Gaussian Splatting (4DGS)~\cite{wu20234d,luiten2023dynamic}, extends this with per-frame dense tracking and novel view synthesis for dynamic scenes by utilizing a lightweight deformation field to model Gaussian motions and shape changes.

\begin{figure*}
    \centering
    \includegraphics[width=\linewidth]{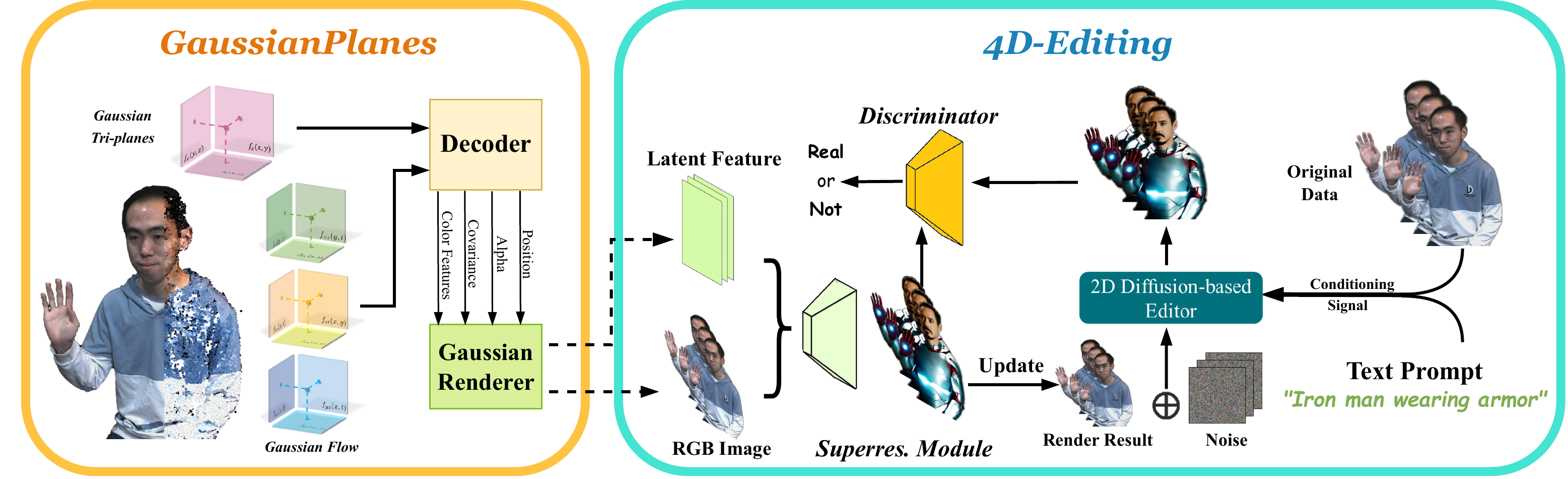}
    \caption{\textbf{\emph{Pipeline of Control4D:}} Our method first utilizes GaussianPlanes to train the implicit representation of a 4D portrait scene, which are then rendered into latent features and RGB images using Gaussian rendering, serving as inputs for the GAN-based generator. Meanwhile, we apply the 2D-diffusion-based editor to edit the dataset with the noisy results and conditions as inputs, leading to updated results that are used as real images while the Superres. Module’s outputs serve as fake images fed into the Discriminator for discrimination. The discriminative results are used to calculate loss, allowing for iterative updates of both the Generator and Discriminator. }
    \label{fig:pipeline}
\end{figure*}

\section{Overview}

To achieve high-quality, efficient, and consistent 4D portrait editing, we first extend the Gaussian Splatting to 4D representation and structure it through a spatial-temporal plane-based decomposition (Sec.~\ref{sec:gaussianplanes}). To address the issue of inconsistencies in edited images generated by diffusion-based editors, we integrate a 4D Editor with GaussianPlanes to effectively mitigate instability and blurring issues and achieve the realism and quality of 4D editing (Sec.~\ref{sec:learn2gen}).
Building on the GaussianPlanes and 4D Editor, we finally introduce several efficient training strategies for 4D editing in the Control4D framework (Sec.~\ref{sec:stage_training}).

As shown in Fig.~\ref{fig:pipeline}, our framework consists of the GaussianPlanes and a 4D generator. Given multi-view videos, we first reconstruct the 4D portrait based on GaussianPlanes. Subsequently, we edit the reconstructed rendering results and latent features through a multi-level generator to obtain the edited results. Simultaneously, we employ an iterative approach to achieve dataset update through a 2D diffusion-based editor, which is a ControlNet~\cite{Zhang22} in practice.  The outputs of the editor serve as real images, while the generator's results function as fake images for the discriminator's input. As the GAN training progresses, we progressively incorporate the generator's outputs to refine the inputs of the 2D diffusion-based editor, facilitating training convergence. Ultimately, the discrimination outcomes are utilized to compute the GAN loss, driving the iterative refinement of both the generator and discriminator. This methodology ensures the efficient and precise realization of 4D editing through our GAN-based framework.

\section{GaussianPlanes}
\label{sec:gaussianplanes}
In this section, we propose an efficient and robust 4D representation GaussianPlanes for 4D portrait editing. The key idea is to structure the discrete points of Gaussian Splatting in both space and time. First, we introduce the spatial tri-plane decomposition, which makes Gaussian Splatting structured in the spatial domain (Sec.~\ref{sec:gaussianplanes:3d}). Following this, we expand Gaussian Splatting into 4D representation and structure the flow of each Gaussian point by performing a temporal-spatial plane-based decomposition (Sec.~\ref{sec:gaussianplanes:4d}).

\subsection{GaussianPlanes in 3D}
\label{sec:gaussianplanes:3d}
Gaussian Splatting is an emerging explicit 3D representation that utilizes a set of Gaussian point clouds to represent 3D scenes. Each point is described with attributes of the center position $\mathbf{x} \in \mathbb{R}^3$, the rotation quaternion $\mathbf{r} \in \mathbb{R}^4$, the scale factor $\mathbf{s} \in \mathbb{R}^3 $, the opacity value $\alpha \in \mathbb{R} $ and the color feature $\mathbf{c} \in \mathbb{R}^3$. 
The rendering process of Gaussian Splatting involves projecting the Gaussian point cloud onto the rendering viewpoint according to camera parameters, followed by rasterization and volume rendering.
Since each point in the Gaussian point cloud is independent and unstructured, noise easily occurs during optimization. To enhance robustness, we propose a spatial tri-plane decomposition to represent the attributes of the Gaussian points. Specifically, we decompose the color $\mathbf{c}_i$, opacity $\alpha_i$ and rotation $\mathbf{r}_i$ of $i$-th Gaussian point into tri-plane features:
\begin{equation}
\begin{split}
    \mathbf{c}_i & = f_{\mathbf{c}}(F_\mathbf{c}^{xy}(x_i, y_i), F_\mathbf{c}^{xz}(x_i, z_i), F_\mathbf{c}^{yz}(y_i, z_i)), \\
    \alpha_i & = f_{\alpha}(F_\alpha^{xy}(x_i, y_i), F_\alpha^{xz}(x_i, z_i), F_\alpha^{yz}(y_i, z_i)), \\ 
    \mathbf{r}_i & = f_{\mathbf{r}}(F_\mathbf{r}^{xy}(x_i, y_i), F_\mathbf{r}^{xz}(x_i, z_i), F_\mathbf{r}^{yz}(yi, z_i)),
\end{split}
\end{equation}
where $F^{xy}, F^{xz}, F^{yz}$ are the decomposed feature planes, and $f$ is an MLP that fuses features to predict specific attributes. In this way, although the Gaussian points remain independent, their attributes are structured and low-rank in spatial space, which helps to reduce noise and improve the robustness of Gaussian Splatting. The scale factor $\mathbf{s}$ and center position $\mathbf{x}$ are not decomposed, as splitting Gaussian points would abruptly halve the scale factor and the center position of each point is used for querying attributes itself.

\subsection{GaussianPlanes in 4D}
\label{sec:gaussianplanes:4d}

To extend Gaussian Splatting for 4D editing, we regard the Gaussian point cloud at the first frame as the canonical space and represent the 4D scene at different times by deforming the canonical Gaussian point cloud. Specifically, we define the flow $\hat{\mathbf{x}}, \hat{\mathbf{r}}$ for both position and rotation attributes of Gaussian points. Then, for time $t$, we move each Gaussian point in the canonical space ($t=0$) with the flow:
\begin{equation}
\begin{split}
    \mathbf{x}_i(t) & = \mathbf{x}_i(0) + \hat{\mathbf{x}}_i(t), \\
    \mathbf{r}_i(t) & = \mathbf{r}_i(0) + \hat{\mathbf{r}}_i(t). \\
\end{split}
\end{equation}
In this way, we enhance temporal consistency since the Gaussian point cloud at all times corresponds to its canonical space. However, the flow of each gaussian point is still discrete and independent. To further structure the flow of Gaussian points in space and time, we adopt spatial-temporal plane-based decomposition proposed by Tensor4D~\cite{Shao22} and decompose the flow attributes of $i$-th point into nine feature planes:
\begin{equation}
\begin{split}
    \hat{\mathbf{x}}_{i}(t) & =f_{\hat{\mathbf{x}}}(x_i,y_i,z_i,t) = \pi_3(\Pi_3(F_{\hat{\mathbf{x}}})), \\
    \hat{\mathbf{r}}_{i}(t) & =f_{\hat{\mathbf{r}}}(x_i,y_i,z_i,t) = \pi_3(\Pi_3(F_{\hat{\mathbf{r}}})),
\end{split}
\end{equation}
where $\Pi_3, \pi_3$ are the hierarchical 4D decomposition in Tensor4D and $F$ represents feature planes. Through spatial tri-planar decomposition and 4D plane-based decomposition, we structure the 4D Gaussian Splatting to enhance its consistency while maintaining efficiency.

\section{4D Editing with GaussianPlanes}
\label{sec:learn2gen}
To solve another challenge raised by diffusion-based editors, we propose a GaussianPlane-based 4D generator to edit 4D scene from the 2D inconsistent editing images with stable optimization. Instead of utilizing direct supervision with the edited images~\cite{Brooks22}, our method learns a continuous generation space via GAN~\cite{goodfellow2020generative} to establish a connection between GaussianPlanes and dynamically edited images. Specifically, we integrate GaussianPlanes with a 2D GAN-based super-resolution module into a 4D generator and learn a generation space from the edited images generated by the diffusion model. Leveraging its generative capabilities, the 4D generator can effectively distill knowledge from the diffusion-based editor and distinguish between the rendering images (fake samples) and edited images (real samples). 
Subsequently, GaussianPlanes can be optimized within a continuous generative space supervised by the discrimination loss. 
With such a learning-to-generate mechanism, our method effectively alleviates blurry effects, resulting in high-fidelity and consistent 4D editing.
In the following, we will introduce 1) integrating GaussianPlanes with GAN for 4D scene generation; and 2) the generation with multi-level guidance.

\subsection{Connecting GAN to GaussianPlanes}
To enable the generative ability of GaussianPlanes and higher rendering resolution, we build a 4D generator by connecting the GaussianPlanes representation with a GAN-based super-resolution module. To this end, we first augment each point in the GaussianPlanes with latent features~\cite{chan2022efficient} as additional attributes.
Here we assume the latent features follow a normal distribution, so in practice we augment the Gaussian attributes with their means $\mathbf{\mu}$ and variances $\mathbf{\sigma}$, thus enabling subsequent sampling.  
We also adopt the same tri-plane decomposition for these latent distribution parameters.  Then we can render a ``distribution parameter map'' to sample the latent features, which will be fed into a super-resolution module $G$.
Meanwhile, we also render an RGB image for auxiliary supervision. 
The distribution map consist of a latent mean map and a latent variance map, 
denoted as $\mathbf{I}_\mathbf{\mu}$ and $\mathbf{I}_\mathbf{\sigma}$, respectively, which capture the mean and variance of latent features. By leveraging this distribution map, we then proceed to sample a latent feature map that will be fed into $G$:
\begin{equation}
\mathbf{I}_l = \mathbf{I}_\mathbf{\mu} + t\mathbf{I}_\mathbf{\sigma}, t \sim N(0, 1).
\end{equation}
Then, we concatenate the rendered RGB images $I_r$ and the latent feature maps $I_l$ and feed them into the super-resolution module to synthesize high-resolution images: 
\begin{equation}
\mathbf{I}_{G} = G(\mathbf{I}_{r}, \mathbf{I}_l).
\end{equation}
As mentioned above, the edited images are temporally inconsistent due to the frame-by-frame editing.
To avoid the discrete and inconsistent issue of direct supervision, our method learns a more continuous generation space via GAN from these edited images.
Specifically, the generated images $\mathbf{I}_G$ are considered as fake samples, while the edited images are regarded as real samples. The GAN loss can be formulated as follows:
\begin{equation}
\begin{split}
L_D & = D(\mathbf{I}_{G}) - D(\mathbf{I}_{ed}) + L_{gp} \\
L_G & = -D(\mathbf{I}_{G}),
\end{split}
\label{equ:gan_loss}
\end{equation}
where $\mathbf{I}_{ed}$ are the edited images generated by diffusion-based editor, $D$ is the discriminator and $L_{gp}$ is the Wasserstein GAN gradient penalty loss~\cite{arjovsky2017wasserstein}. 

\begin{figure}
    \centering
    \includegraphics[width=\linewidth]{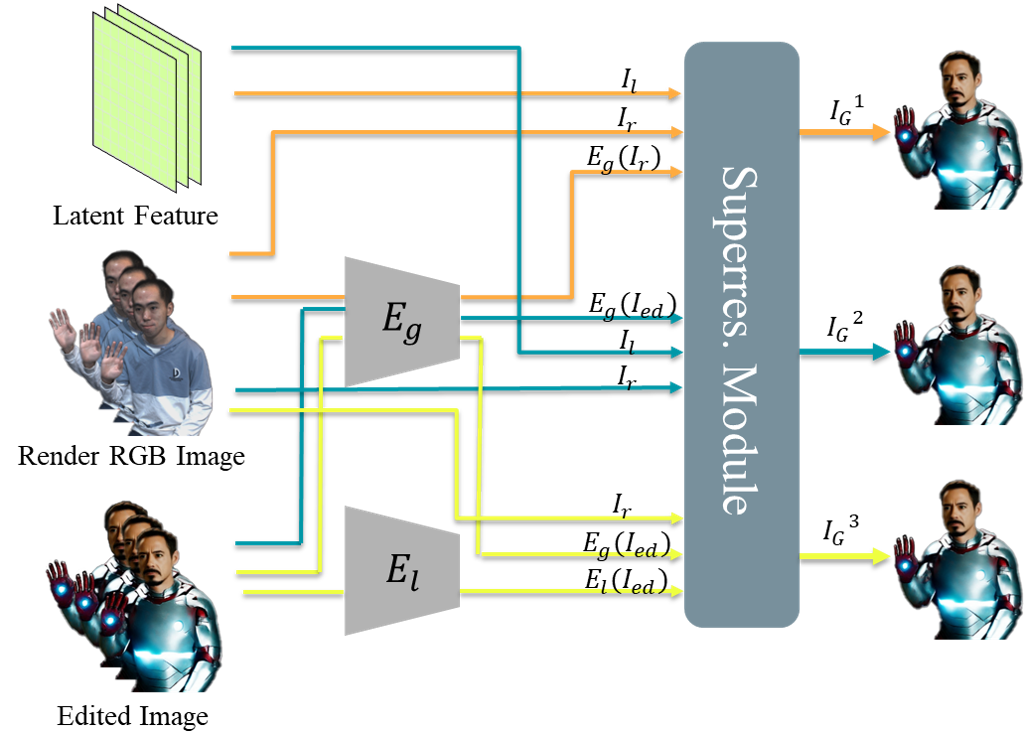}
    \vspace{-8mm}
    \caption{\textbf{\emph{Illustration of the Generation with Multi-level Guidance:}} we propose a three-level image generation process to balance the generator training, where $E_g$ denotes for the global encoder and $E_l$ denotes for the local encoder.}
    \label{fig:multilevel}
\end{figure}

\begin{figure*}
    \centering
    \includegraphics[width=\linewidth]{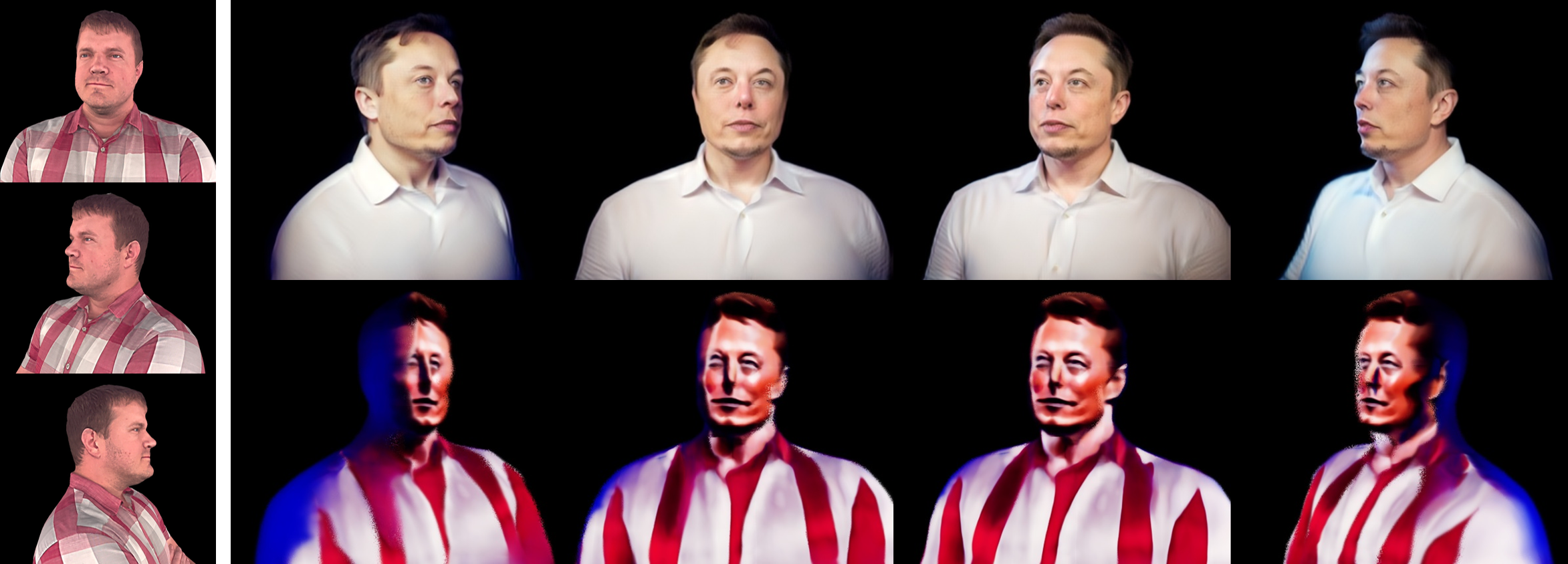}
    \caption{\textbf{\textit{Qualitative comparisons with Instruct-NeRF2NeRF(static):}} In a static scenario, given the prompt “Turn him into Elon Musk”, train the model to converge and we can see that, on the same dataset, our method (the top row) produces highly realistic renderings of human portraits, while instruct nerf2nerf exhibits lower levels of realism and consistency, along with unexpected distortions in facial features. }
    \label{fig:static_cp}
\end{figure*}

\subsection{Multi-level Generation with Guidance}
When training GAN with the loss in Eqn.~\ref{equ:gan_loss}, we observe that the learning process often suffers from mode collapse issue.
This may be caused by the fact that there is a limited number of edited images, and it is easy for the discriminator to learn how to distinguish between different sources of samples.
To stabilize the learning process, we propose to extract multi-level information from the edited images and use these global and local cues to guide the learning of the generator.
As shown in Fig.~\ref{fig:multilevel}, during the training process, we construct two networks—global encoder $E_g$ and local encoder $E_l$—to extract the global code and local feature maps of the edited image $\mathbf{I}_{ed}$, respectively. With these conditions as additional inputs, our generator can synthesize images on three levels:
\begin{equation}
\begin{split}
    \mathbf{I}_{G}^1 & = G(\mathbf{I}_{r}, \mathbf{I}_{l}, E_g(\mathbf{I}_{r})) \\
    \mathbf{I}_{G}^2 & = G(\mathbf{I}_{r}, \mathbf{I}_{l}, E_g(\mathbf{I}_{ed})) \\ 
    \mathbf{I}_{G}^3 & = G(\mathbf{I}_{r}, E_l(\mathbf{I}_{ed}), E_g(\mathbf{I}_{ed}))
\end{split}
\end{equation}
Throughout the progression from level 1 to level 3, the generator produces images that gradually approach real edited images:
\begin{itemize}[leftmargin=*]
    \item At level 1, the generator directly synthesis images based on Tensor4D.
    \item At level 2, global information from the real edited images is introduced as conditions, guiding the generator to produce results consistent with the overall style of the real images. 
    \item At level 3, both the global and local information from the real edited images is used as conditions, enabling the network to generate images that exhibit consistency in both the overall pattern and finer details with the real edited images. 
\end{itemize}
To facilitate training, we also utilize different losses at different levels:
\begin{equation}
\begin{split}
    L_1 & = -D(\mathbf{I}_{G}^1) \\
    L_2 & = -D(\mathbf{I}_{G}^2) + L_P(\mathbf{I}_{G}^2, \mathbf{I}_{ed}) \\ 
    L_3 & = -D(\mathbf{I}_{G}^3) + L_P(\mathbf{I}_{G}^3, I_{ed}) + \|\mathbf{I}_{G}^3 - \mathbf{I}_{ed} \|_1
\end{split}
\end{equation}
Level 1 employs the original GAN loss. At level 2, a perceptual loss is introduced as an additional constraint to enforce consistency in the global style. At level 3, the loss function simultaneously incorporates L1 loss, perceptual loss, and GAN loss as penalties, as the consistency in details and global style is desired. This multi-level information guides the generator to converge progressively towards the generation space of the diffusion model, improving training stability in single scenarios and accelerating convergence compared to the original GAN training process.

\subsection{Training Strategy}
\label{sec:stage_training}
To address the high iterative optimization cost associated with using the diffusion-based editor, we propose several strategies to further improve the efficiency of 4D editing.

\noindent \textbf{Staged Training Strategy.}
We adopt a staged training strategy that facilitates convergence. First, we fix the flow in the static stage and focus solely on editing the canonical space. This simplifies the editing process from 4D to 3D static editing, resulting in faster convergence. Once the editing of the canonical space has converged, we proceed to train GaussianPlanes across the entire 4D sequences. We also adopt a smaller noise timestep $t \in U(0.02, 0.6)$ for the diffusion-based editor in the dynamic stage since most of the editing effect is done in the static stage.

\noindent \textbf{Batch-based Dataset Update.}
To improve the editing consistency across different images, instead of editing a single image per iteration like InstructN2N, we group several images as a batch and edit them simultaneously. In editing each batch, we incorporate an attention module~\cite{guo2023animatediff} for multi-frame image generation into our diffusion-based editor to capture the temporal-spatial correspondences and thereby improve editing consistency.

\begin{figure*}[t]
    \centering
    \includegraphics[width=\linewidth]{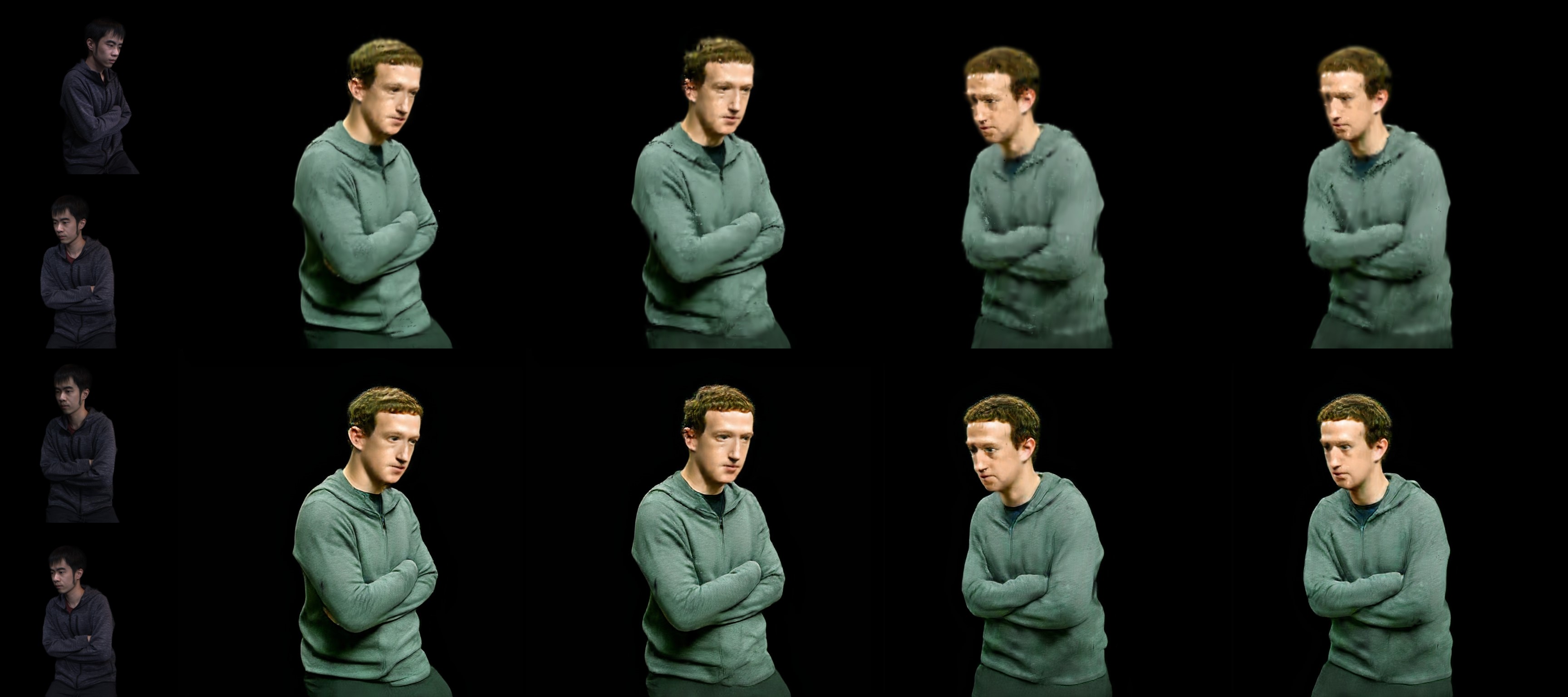}
    \caption{\textbf{\textit{Qualitative comparisons with baseline(dynamic):}} In a dynamic scenario, given the prompt “Mark Zuckerberg”, compared to the baseline result (the first row) that only employs the dataset update (DU) method, our proposed approach (with the addition of GAN, the second row) demonstrates higher levels of realism and consistency in our rendered results.  }
    \label{fig:dynamic_cp}
\end{figure*}

\begin{figure*}[ht]
    \centering
    \includegraphics[width=\linewidth]{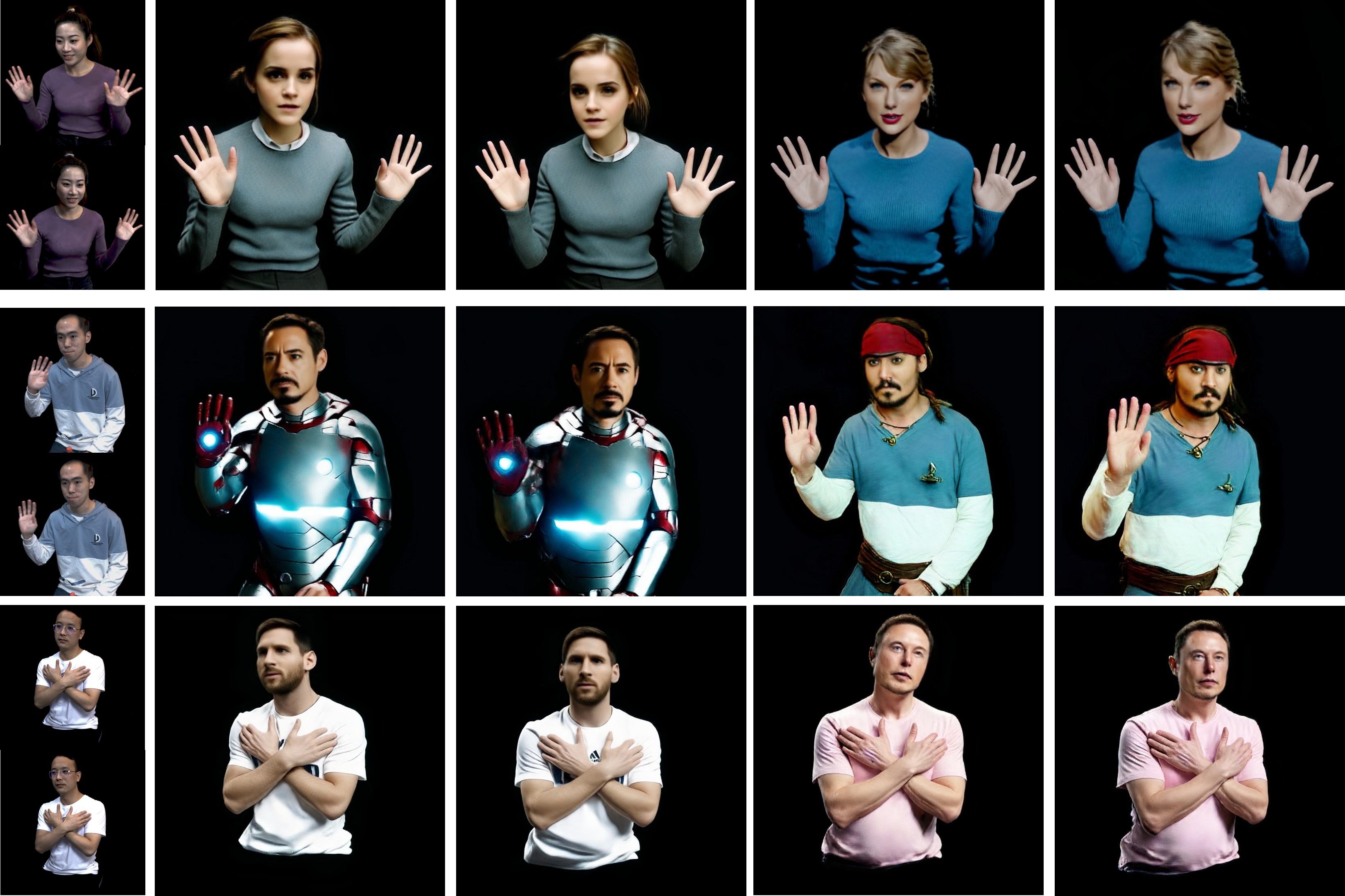}
    \caption{Qualitative results on Tensor4D dataset. Our method produces coherent and realistic outcomes that maintain high fidelity and accurately preserve intricate details throughout the dynamic editing procedure. The prompts we use are "Emma Watson", "Taylor Swift", "Iron Man wearing armor", "Captain Jack Sparrow",  "Lionel Messi" and "Elon Musk".}
    \label{fig:result}
\end{figure*}

\section{Experiment}
We primarily conduct experiments on the dynamic Tensor4D dataset, which captures dynamic half-body human videos by four sparsely positioned, fixed RGB cameras. The calibration is performed using a checkerboard. Each data sample captures a diverse range of human motions in 1-2 minute duration. For our experiments, we extract 2-second segments, consisting of 50 frames, from the full-length videos for 4D reconstruction and editing. Furthermore, to showcase the capabilities of our method in 360-degree scenes, we also select scanned human models from Twindom~\cite{yu2021function4d} dataset for additional evaluation. Please refer to the suppl. for more experiment details.

\subsection{Qualitative Evaluation}
\subsubsection{Static scene} 
Since the task of 4D editing with text has not been addressed in previous works, we first conduct evaluation on static scene in order to validate the efficiency of our proposed methods. 
To validate the efficiency of our proposed GAN, we first conduct a comparison between NeRF+GAN and instruct-NeRF2NeRF under static scenes. We select some human models from the Twindom dataset and sampled 180 viewpoints randomly within a 360-degree range to render images. Subsequently, we evaluate NeRF+GAN and instruct-NeRF2NeRF for editing with prompt "Turn him into Elon Musk". In Fig.~\ref{fig:static_cp}, we present the results after 50,000 iterations of training. Observing the results, it is evident that our GAN can generate images of high quality, exhibiting rich detail and enhanced realism. In contrast, the instruct-NeRF2NeRF outputs appear smoother, with some issues observed in the blending of side views. This comparison highlights the significant advantage of our GAN in terms of editing capabilities.

\begin{table}[t]
    \centering
    \begin{tabular}{p{3.5cm}|cc}
        Method &  FID$\downarrow$  & CLIP Similarity $\uparrow$ \\ \hline
        \textbf{Static scene} &   &  \\
        InstructPix2Pix & - & 0.3089\\
        ControlNet & - & 0.3313 \\
        InstructNeRF2NeRF & 126.3 & 0.2989 \\
        Tensor4D & 118.7 & 0.3316 \\
        Tensor4D+GAN & 27.81 & \textbf{0.3334} \\
        GaussianPlanes & 49.32 & 0.3301 \\ 
        Control4D (Ours) & \textbf{14.11} & 0.3323 \\ \hline
        \textbf{Dynamic scene}  &   & \\
        ControlNet & - & 0.3185\\
        Tensor4D & 155.6  & 0.3144 \\
        Tensor4D+GAN & 47.39 & 0.3178 \\ 
        GaussianPlanes & 67.58 & 0.3175 \\ 
        Control4D (Ours) & \textbf{18.59} & \textbf{0.3192} \\ 
    \end{tabular}
    \caption{Quantitative Comparisons on static and dynamic scenes.}
    \label{tab:quan_expr}
\end{table}

\subsubsection{Dynamic scene} 
In dynamic scenarios, we compare our proposed method and the baseline method that only utilize GaussianPlanes, and the results are presented in Figure~\ref{fig:dynamic_cp}. We also present the results of different individuals engaged in various actions, which can be referenced in Figure~\ref{fig:result}.
In the baseline approach, where GAN-based generation is not utilized, GaussianPlanes is directly tasked with fitting a dynamically changing editing dataset in both space and time. This direct fitting process often leads to the optimization of smooth results that may lack consistency and high-quality details.
Our proposed method incorporates GAN-based generation, leveraging the GAN to learn a more continuous 4D generation space. This allows us to leverage the smooth supervisory signals for optimization. Thus, our method generates consistent and high-quality results that exhibit improved fidelity and capture finer details in the dynamic editing process.
The comparison between the baseline approach and our method demonstrates the effectiveness of our proposed 4D generator in enhancing the overall quality and consistency of the generated results.

\subsection{Quantitative Experiment}
We conducted quantitative experiments in 5 static and 4 dynamic scenarios. The results are presented in Tab.~\ref{tab:quan_expr}. First, we compare the diffusion-based editor including Instructpix2pix~\cite{Brooks22} and ControlNet~\cite{Zhang22} in the context of portrait editing. ControlNet exhibited better consistency between the subject and the editing prompt than Instructpix2pix. We further compared our method, Control4D, with the baseline approaches including Tensor4D, Tensor4D+GAN, and GaussianPlanes to validate the efficiency of our proposed representation and GAN. We evaluated the Fréchet Inception Distance (FID) metric~\cite{heusel2017gans} between the edited dataset and generated images. We also compute CLIP cosine similarity~\cite{CLIP} between the generated images and text. Compared with Tensor4D and Tensor4D+GAN, our method achieves superior performance, which demonstrates the efficiency of GaussianPlanes. The results also reveals that our method outperforms the baseline and InstructNeRF2NeRF~\cite{Instruct-NeRF2NeRF} significantly, demonstrating the effectiveness of our proposed 4D editing pipeline.
\section{Conclusions}
In conclusion, Control4D is a novel approach for efficient, high-fidelity and temporally consistent editing in dynamic 4D scenes. It utilizes an efficient 4D representation GaussianPlanes and a 2D diffusion-based editor. By utilizing plane-based decomposition to struct Gaussian Splatting, GaussianPlanes ensure both efficiency and robustness for 4D editing. To tackle with the inconsistency caused by diffusion-based editor, Control4D leverages a GAN to generate from the editor, avoiding direct supervision. 
Experimental results demonstrate Control4D's effectiveness in achieving photo-realistic and consistent 4D editing, surpassing previous approaches in real-world scenarios. It represents a significant advancement in text-based image editing, particularly for dynamic scenes.

\textbf{Limitations.}
Due to utilizing a canonical Gaussian point clouds with flow representation, our approach relies on learning flow within the 4D scenes to exhibit simplicity and smoothness. This poses challenges for our method in effectively handling rapid and extensive non-rigid movements. Furthermore, our method is constrained by ControlNet, which limits the granularity of edits to a coarse level. Consequently, it is unable to perform precise expression or action edits. Our method also requires iterative optimizations for the editing process and cannot be accomplished in a single step.

{
    \small
    \bibliographystyle{ieee_fullname}
    \bibliography{egbib}
}

\appendix
\clearpage
\setcounter{page}{1}
\maketitlesupplementary

\section{Implementation Details}
\subsection{GaussianPlanes}
\noindent \textbf{Spatial triplane decomposition.} In 3D spatial space, we decompose the attributes of each point in the canonical Gaussian point cloud, including color, opacity, latent feature, and rotation. The decomposition of each attribute utilizes three corresponding feature planes. We employ a HashGrid~\cite{muller2022instant} to represent each feature surface, with a hierarchy of resolutions at 16 levels, where the scale of each level is 1.3 times that of the preceding level. Each level contains 2 feature channels, and the encoded results are mapped to the corresponding attributes through a 256-unit MLP network.

\noindent \textbf{4D flow decomposition.}
For the 4D flow, we have implemented a hierarchical decomposition using Tensor4D~\cite{Shao22}. In our approach, we decompose the flow of each point's position and rotation at each moment within the Gaussian point cloud. This decomposition employs 9 feature planes. Each feature plane is represented using a HashGrid consistent with the spatial feature planes. Subsequently, the encoded results are first fused individually for the corresponding three planes using three 256-unit MLPs, and then a final attribute output is produced through another 256-unit MLP.

\subsection{4D Generator}

\noindent \textbf{Network Structure.} We adopt a network architecture similar to pix2pixHD~\cite{wang2018high} for the GAN's generator and discriminator. In our generator, we introduce some modifications: the input features comprise both RGB and latent features, which are concatenated together as the input. Additionally, in the intermediate layers, we concatenate the feature with its global feature code. The architecture includes three downsample layers, three middle blocks, and five upsample layers, thereby achieving a 4x super-resolution in the final output. The number of the base feature channel in the network is 32. The input for the 4D generator is at a resolution of 256, and it outputs images at a resolution of 1024. As for the discriminator, we utilize the same architecture as the pix2pixHD discriminator.

\begin{figure*}[t!]
    \centering
\includegraphics[width=\linewidth]{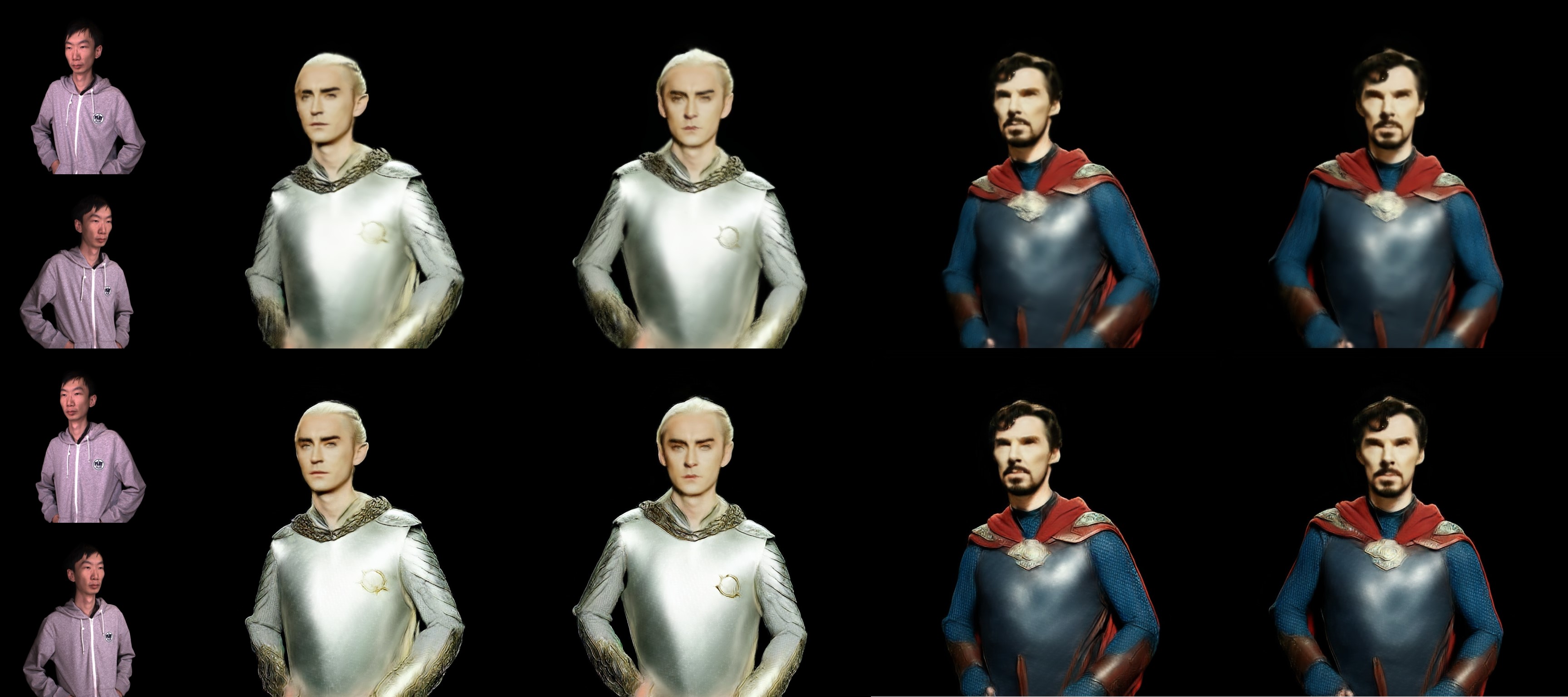}
    \caption{Ablation study of 4D generator. First row: Results utilizing only GaussianPlanes, second row: Results achieved by combining 4D generator. The prompts used here are "Elf King" and "Doctor Strange".}
    \label{fig:ablation-4D-generator}
\end{figure*}

\begin{figure*}[t!]
    \centering
\includegraphics[width=\linewidth]{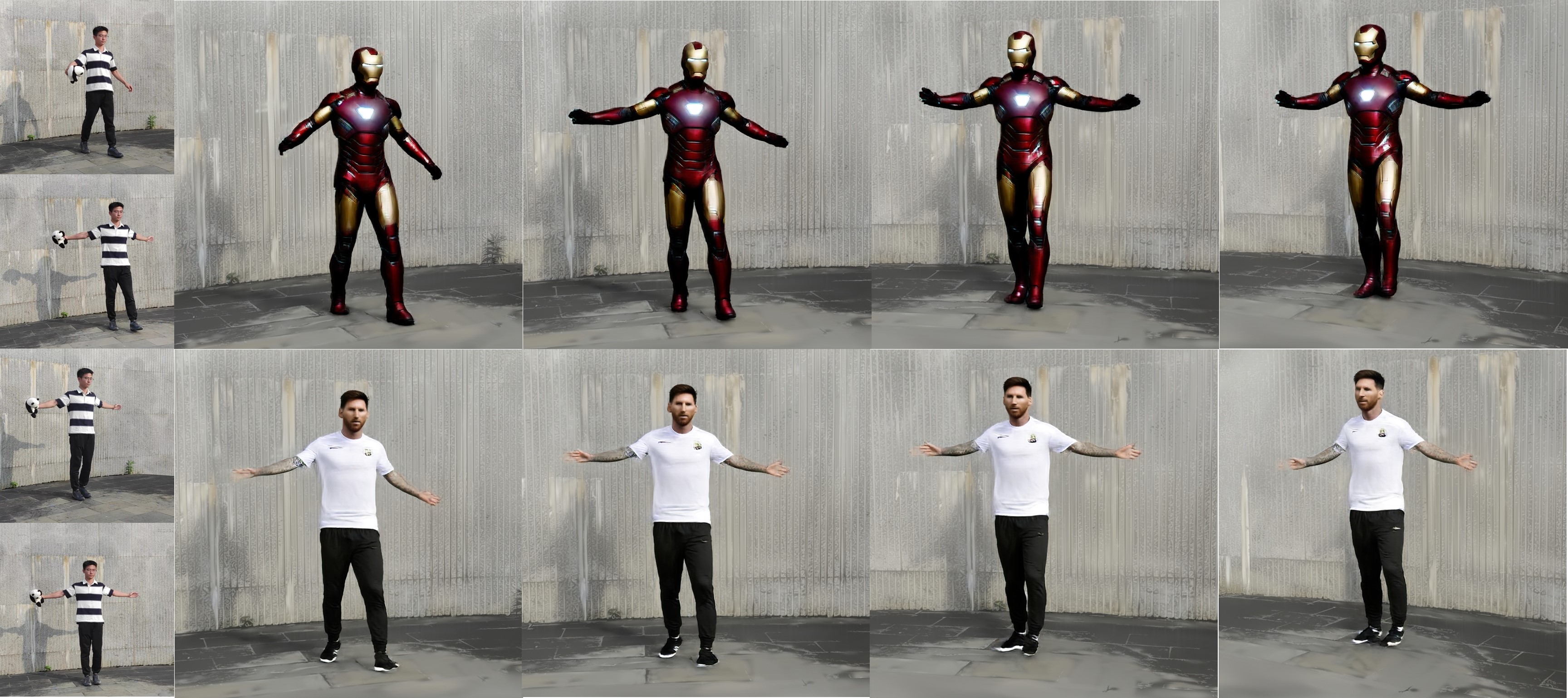}
    \caption{Control4D results on ENeRF dataset. The prompts are "Iron Man" and "Lionel Messi”.}
    \label{fig:enerf-dataset}
\end{figure*}

\begin{figure*}[t!]
    \centering
\includegraphics[width=\linewidth]{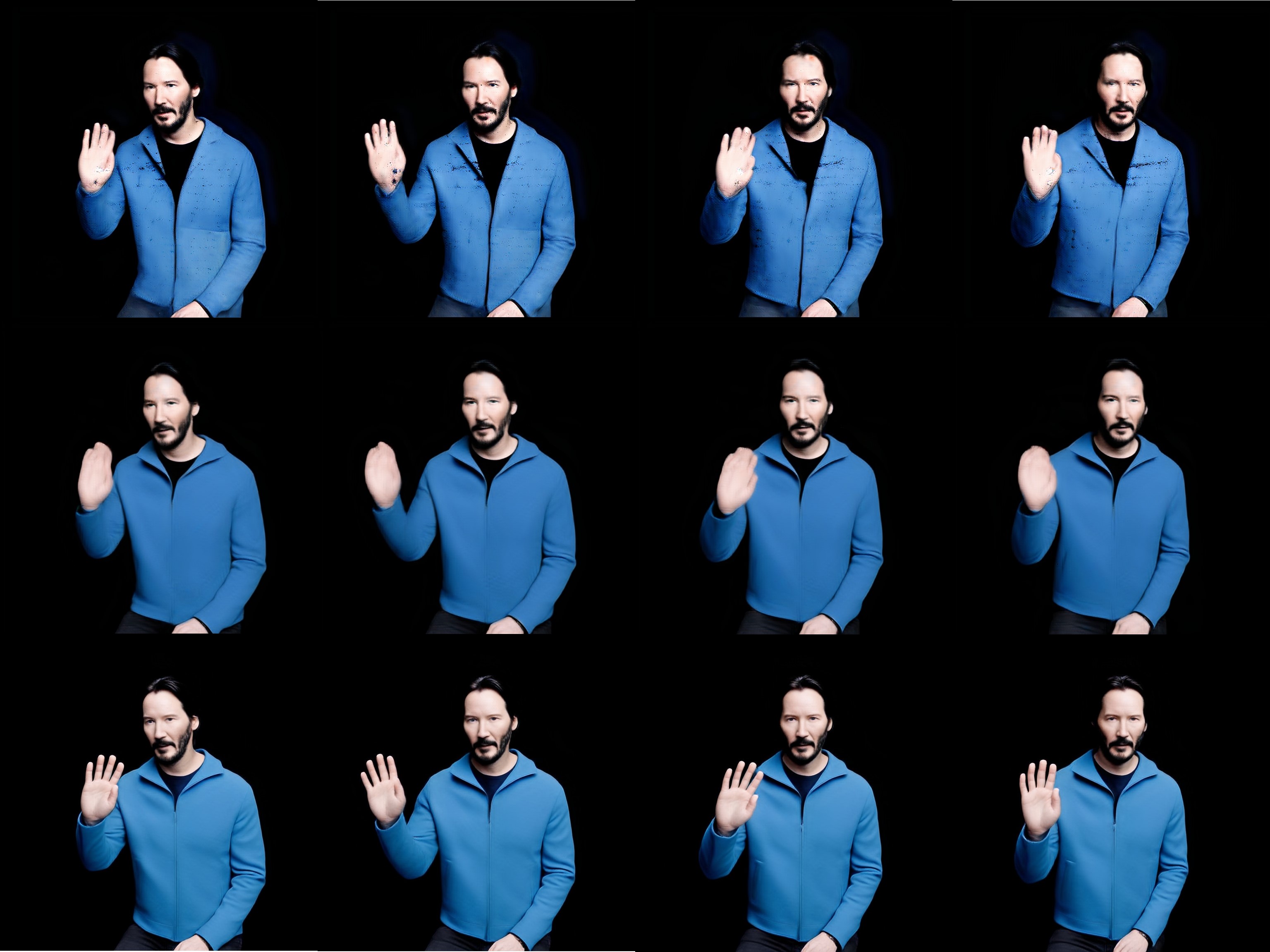}
    \caption{Ablation Study of GaussianPlanes. First row: w/o spatial triplane decomposition. Second row: w/o 4D flow decomposition. Third row: Control4D results.}
    \label{fig:ablation-gaussianplanes}
\end{figure*}

\begin{figure*}[t!]
    \centering
\includegraphics[width=\linewidth]{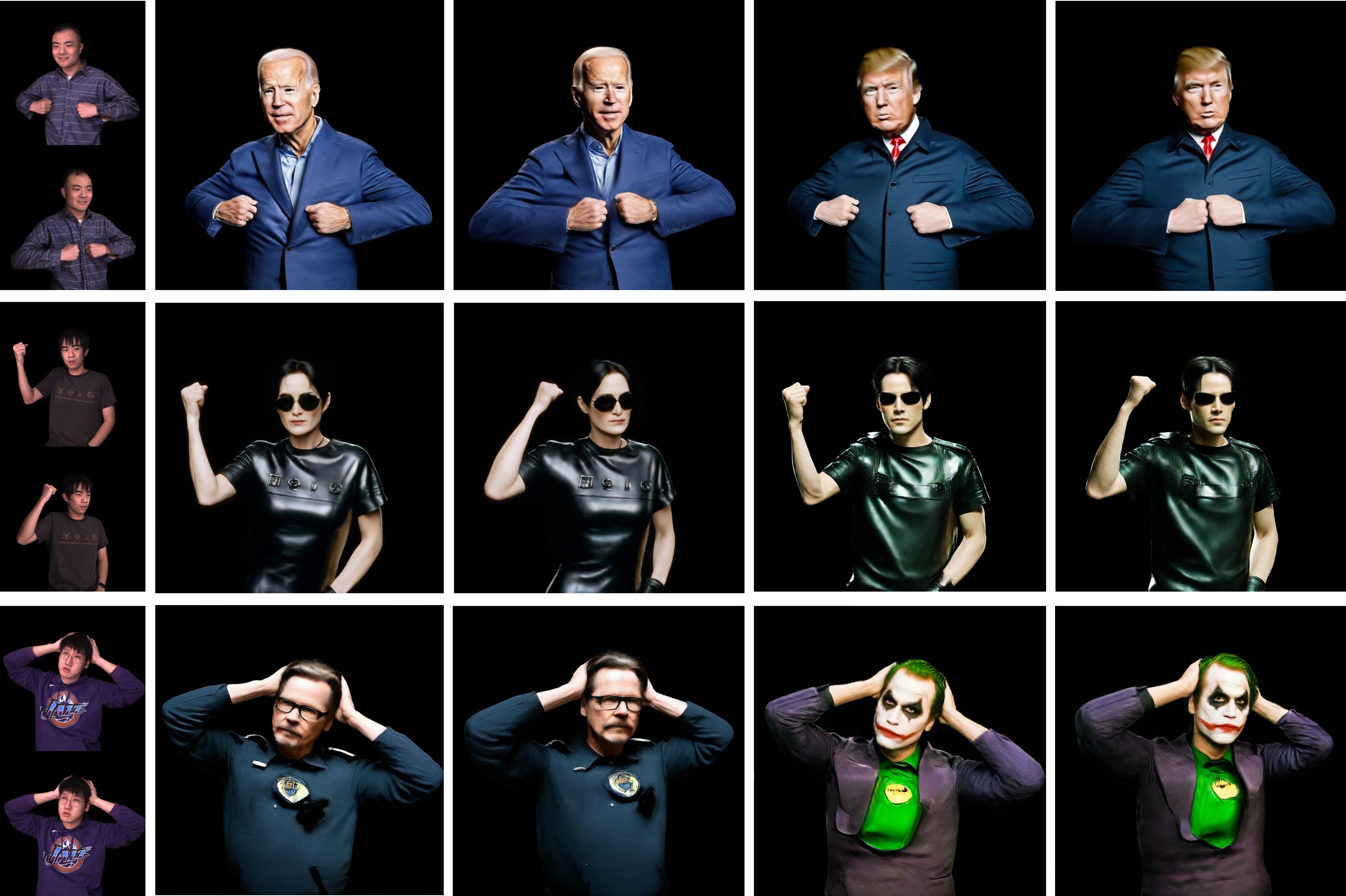}
    \caption{More Control4D results on Tensor4D dataset. The prompts are "Joe Biden wearing suit", "Donald Trump wearing suit", "Trinity in The Matrix", "Neo in The Matrix", "James Gordon in Batman", and "Joker in Batman".}
    \label{fig:t4d-result}
\end{figure*}

\begin{figure}[t!]
    \centering
\includegraphics[width=\linewidth]{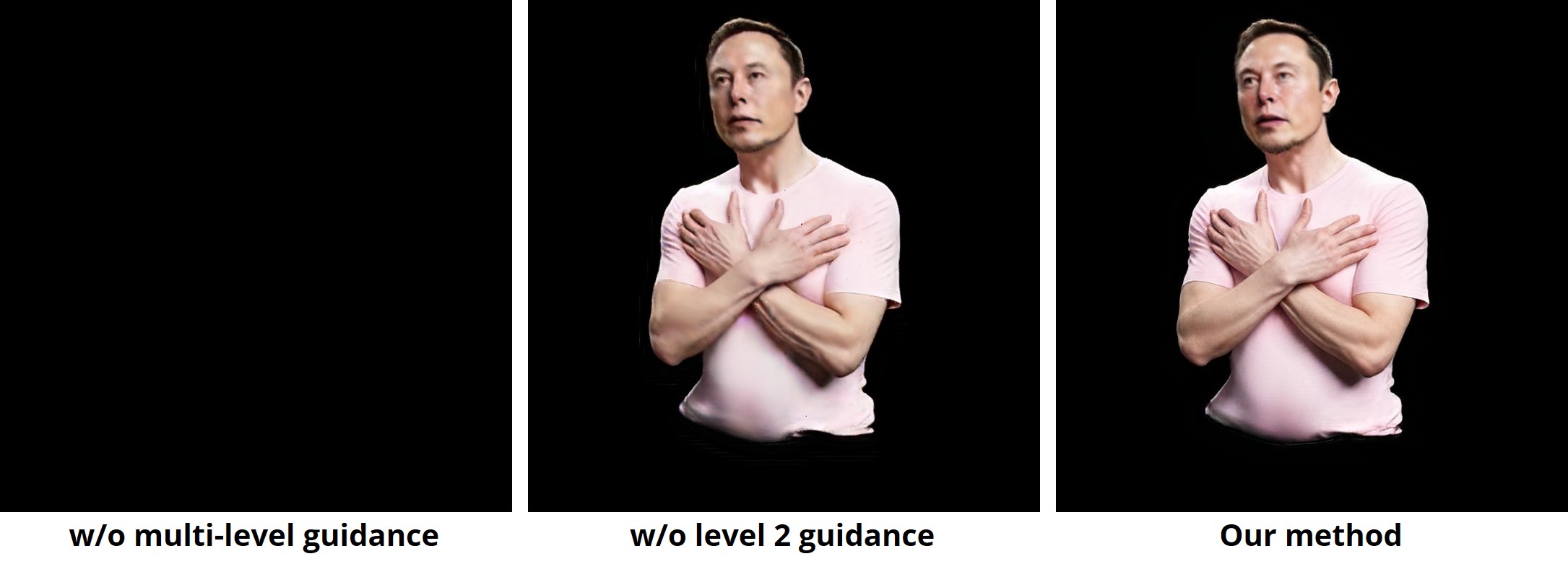}
    \caption{Ablation study of multi-level guidance.}
    \label{fig:multi-level}
\end{figure}

\begin{figure*}[t!]
    \centering
\includegraphics[width=\linewidth]{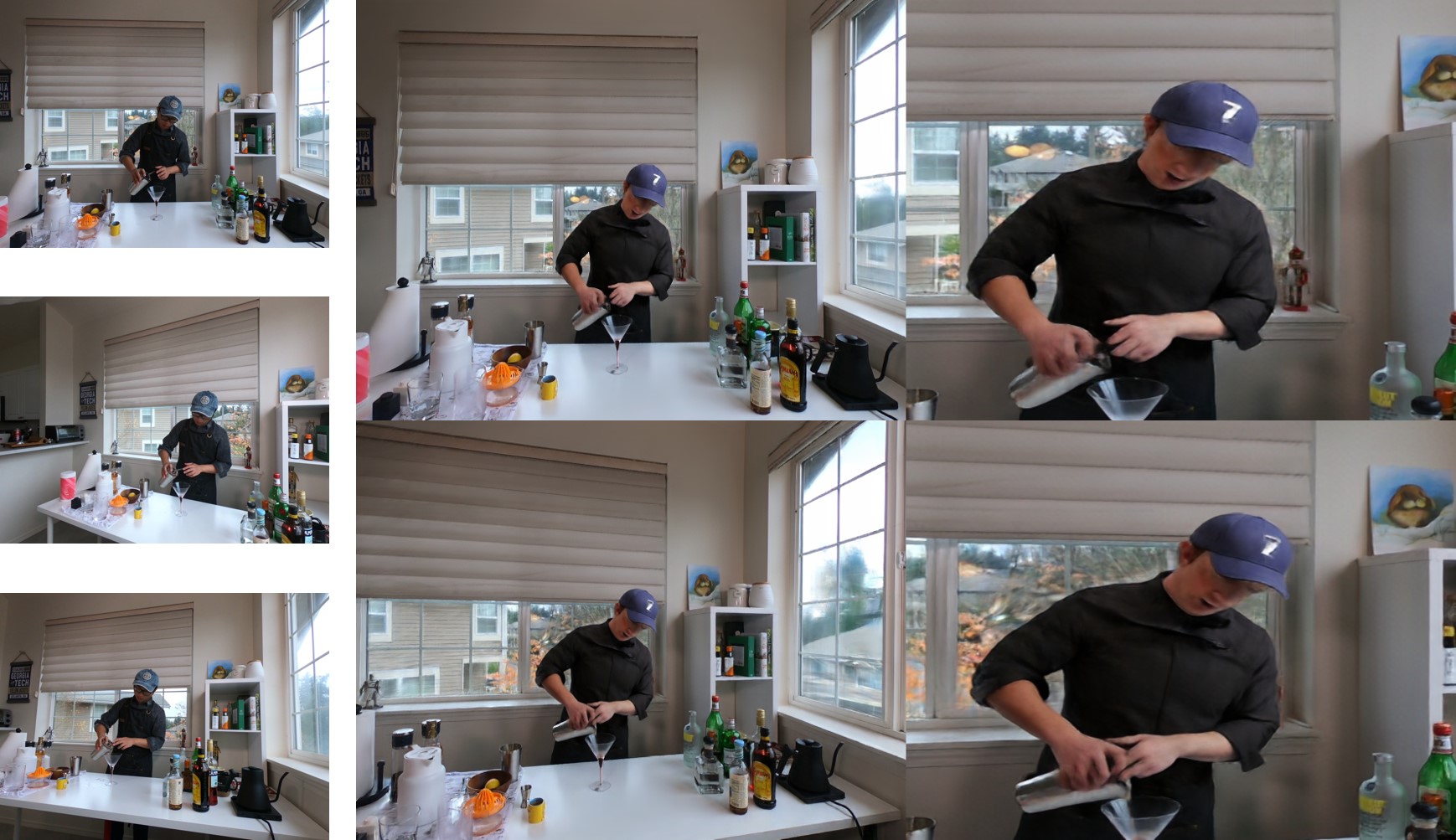}
    \caption{Control4D result on neural 3D video dataset. The prompt is "Mark Zuckerberg". }
    \label{fig:n3d-result}
\end{figure*}

\begin{figure*}[t!]
    \centering
\includegraphics[width=\linewidth]{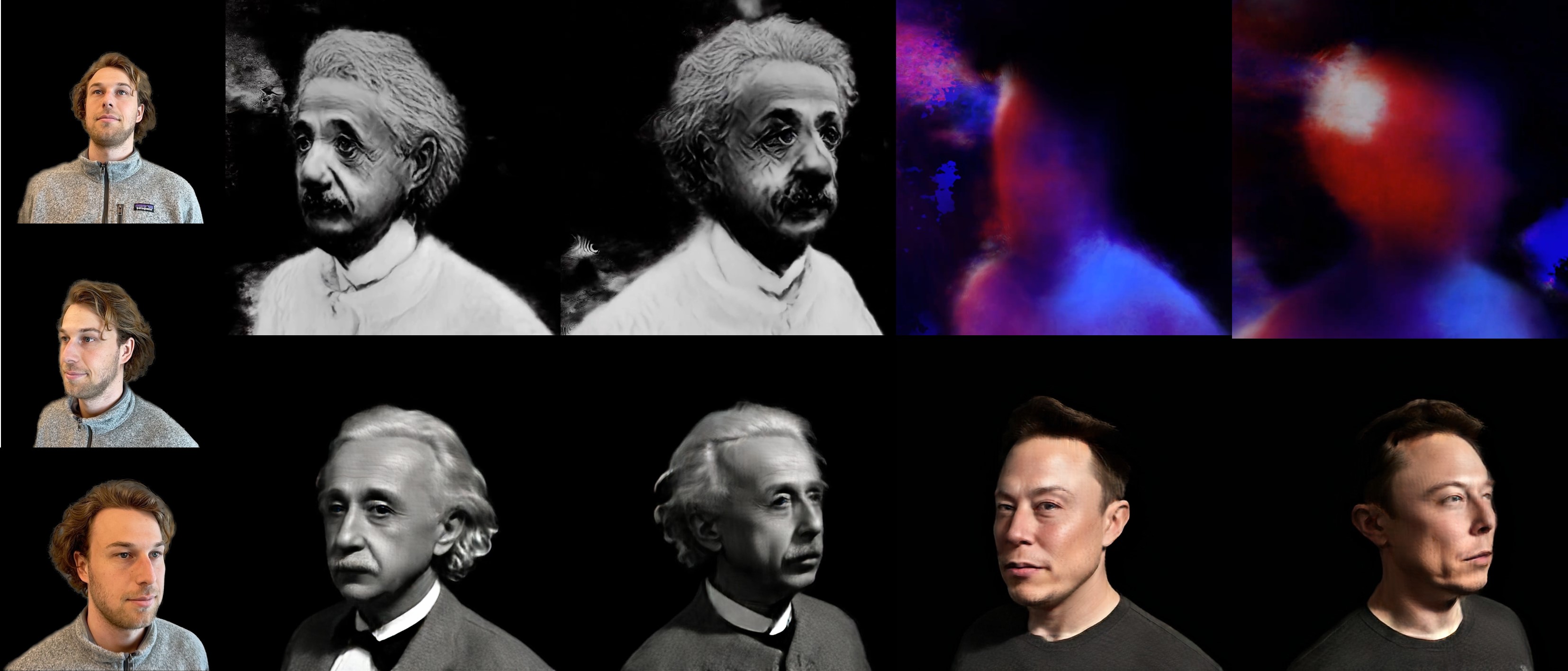}
    \caption{Comparison with InstructNeRF2NeRF on InstructNeRF2NeRF dataset. First row: InstructNeRF2NeRF results with prompts "Turn him into Albert Einstein" and "Turn him into Elon Musk". Second row: Control4D results with prompts "Albert Einstein" and "Elon Musk". }
    \label{fig:compare-in2n}
\end{figure*}

\noindent \textbf{Global Encoder.} We utilize MobileNet~\cite{howard2017mobilenets} to extract the global code. Initially, the image is resized to a resolution of 224, followed by feature extraction through MobileNet's layers. We maps the final feature of MobileNet to a 64-dimensional global feature code.

\noindent \textbf{Local Encoder.} We employ an encoder similar to the VAE encoder used in Stable Diffusion~\cite{Rombach22} as our local encoder. Our local encoder compresses the original image to a quarter of its original size through two downsample layers, and the number of "z\_channels" is set to 4. The base channel number of our network is 32.

\subsection{Diffusion-based Editor} We utilize ControlNet~\cite{Zhang22} as our diffusion-based editor. To achieve better control effects, we employ both normal and OpenPose as control signals. The control strength for normal is set at 0.5, while for OpenPose, it is 1.0. Additionally, we use the RealisticVision~\cite{realistic}, an SD1.5 model, to obtain more realistic editing effects. Additionally, before feeding the images into ControlNet, we resize the 1024-resolution images down to a resolution of 512.

\subsection{4D Reconstruction based on GaussianPlanes}
We initially reconstruct the 4D scenes using GaussianPlanes. Our experiments primarily focus on the Tensor4D dataset, and we also showcase some results in challenging scenes, including those from Neural3DVideo~\cite{Li21}, ENeRF~\cite{lin2022efficient} and InstructNeRF2NeRF~\cite{Instruct-NeRF2NeRF}. For the Tensor4D dataset, we employ a Gaussian sphere for initialization, with a point cloud size of 5,000 and a radius of 1. For the ENeRF, Neural3DVideo and InstructNeRF2NeRF datasets, we utilize the point cloud from the first frame of COLMAP~\cite{schoenberger2016sfm} as the initialization.

During the training process, to ensure the stability of the canonical Gaussian point cloud, we adopt a weighted strategy for selecting training frames. There is a 50\% probability that we choose all frames from the first moment and a 50\% probability that we randomly select frames from other moments. This approach is designed to balance the representation of the initial frame with the dynamic aspects of the remaining video content.
Simultaneously, we are also training the 4D generator in preparation for 4D editing. 

During the training process, the learning rate for the point cloud positions linearly decays from 0.00016 to 0.0000016. The learning rates for scaling, color, opacity, and rotation are set at 0.005, while the learning rate for flow is 0.00025. The learning rate for the 4D generator is 0.001. The gradient threshold for splitting is set to 0.0002, and the interval of densification and pruning is 200. We employ L1 loss to train GaussianPlanes, with a weight of 1.0. For training the generator, we use L1 loss, perceptual loss, and GAN loss, with weights of 1.0, 1.0, and 0.01, respectively. The discriminator is trained using GAN loss and gradient penalty regularization, with respective weights of 1.0 and 0.01.
\subsection{4D Editing Process}
During the 4D editing process, we utilize two GPUs (RTX3090) for training. One GPU is dedicated to running edits for each image, while the other GPU is tasked with running GaussianPlanes and rendering images with the 4D generator. These two processes are executed in parallel. For complex multi-camera 4D scenes, including Neural3DVideo and ENeRF, we do not edit using all cameras. Instead, we use images from all cameras at the first moment and randomly select images from four cameras at other moments to form the dataset.

The first 1000 steps of our training are for static editing, followed by 4000 steps for dynamic editing. During static editing, the noise added to the diffusion-based editor is $U(0.02, 0.98)$, which is reduced to $U(0.02, 0.6)$ for dynamic editing. The steps of diffusion model is set to $20$ and we use the DDIM scheduler. To enhance robustness, we lower the learning rates during the editing process. Specifically, the learning rate for point positions is 0.000016, while the learning rates for scaling, color, opacity, and rotation remain at 0.005, and the learning rate for flow is 0.0001. The 4D generator's learning rate is set at 0.0001. In the editing process, we don't split or prune Gaussian points. In the multi-level guidance, the probability of selecting each level is equal. The weights of the various losses in 4D editing remain consistent with those in the reconstruction process.

\section{More Comparisons}
We conducted further comparisons with InstructNeRF2NeRF on their dataset. As shown in Fig.~\ref{fig:compare-in2n}, our method noticeably surpasses InstructNeRF2NeRF in terms of realism and quality. Additionally, our optimization process is extremely fast, completing editing tasks in just 5 minutes, whereas InstructNeRF2NeRF requires at least about 5 hours. This makes our method 60 times more efficient than InstructNeRF2NeRF.

\section{More Ablation Study}
\noindent \textbf{GaussianPlanes.} We conducted more ablation studies on GaussianPlanes. As shown in Fig.~\ref{fig:ablation-gaussianplanes}, when the spatial triplane decomposition is not used, the results exhibit significant noise. Without the decomposition of flow, the edited results become noticeably blurred, with evident occurrences of jittering, which can be clearly observed in the supplementary video. This demonstrates that our proposed plane decomposition method makes Gaussian Splatting more structured and significantly enhances its robustness. 

\noindent \textbf{4D generator.} More ablation experiments were conducted on our proposed 4D generator. As illustrated in Fig.~\ref{fig:ablation-4D-generator}, without the use of the generator and relying solely on GaussianPlanes, the images noticeably lose many high-quality details and appear blurry. This validates the role of our 4D generator in enhancing quality.

\noindent \textbf{Multi-level guidance.}
 Further ablation studies were performed on multi-level guidance. As shown in Fig.~\ref{fig:multi-level}, when only GAN loss is used, mode collapse occurs easily due to the small size of the dataset. When only the first and third levels are used, the results become blurred. This indicates that our progressive guidance strategy improves the stability of the GAN and gradually enhances the quality of the rendered images.
 
\section{More Results}
Our method is applicable not only to the editing of half-body and heads but also to complex 4D scenes and full-body human editing. More results are illustrated in Fig.~\ref{fig:n3d-result}, \ref{fig:t4d-result}, and \ref{fig:enerf-dataset}. For dynamic editing effects, please refer to our supplementary video.

\section{Social Impact}
The primary goal of our method is to provide users with an advanced tool for dynamic human editing in complex 4D scenes. While our approach enables intricate editing of full-body humans and facilitates creative expression in digital environments, it also raises concerns about potential misuse, such as creating deceptive or misleading content. This challenge is not exclusive to our method but is a common issue across various generative modeling techniques. Additionally, in line with ethical considerations, our approach underscores the importance of diversity, including aspects of gender, race, and cultural representation. It is crucial for ongoing and future research in generative modeling to continuously engage with and reevaluate these ethical considerations to ensure responsible use and positive societal impact.

\end{document}